\documentclass{article}
\usepackage{spconf,amsmath,graphicx,subcaption}
\usepackage{xcolor}
\usepackage[breaklinks]{hyperref}
\usepackage{caption}
\usepackage{floatrow}
\usepackage{enumitem}
\setlist{nosep, leftmargin=14pt}

\usepackage{mwe} 
\usepackage{svg} 
\usepackage{pifont} 

\newcommand{\cmark}{\ding{51}}%
\newcommand{\xmark}{\ding{55}}%

\newcommand{\vqamed}{{\sc vqa}-{\sc m}ed 2019\xspace}
\newcommand{\mmbert}{{\sc mmbert}\xspace}
\newcommand{\roco}{{\sc roco}\xspace}
\newcommand{\bleu}{{\sc bleu}\xspace}
\DeclareUnicodeCharacter{0308}{ }

\title{MMBERT: M\MakeLowercase{ultimodal} BERT P\MakeLowercase{retraining}  \MakeLowercase{for} i\MakeLowercase{mproved} M\MakeLowercase{edical} VQA}

\name{Yash Khare$^{\star \dagger}$ \quad \quad Viraj Bagal$^{\star \ddagger}$\quad \quad Minesh Mathew$^{\dagger}$
\thanks{$^{\star}$ Equal Contribution. Order decided by coin toss. Work done during an internship at IIIT Hyderabad.}}
\secondlinename{Adithi Devi$^{\dagger \dagger}$\quad U Deva Priyakumar$^{\dagger}$ \quad CV Jawahar$^{\dagger}$}
\address{$^{\dagger}$ IIIT Hyderabad, India \quad  $^{\ddagger}$IISER Pune, India \quad $^{\dagger \dagger}$Osmania Medical College, India}

\begin{document}

\maketitle

\begin{abstract}

Images in the medical domain are fundamentally different from the general domain images. Consequently, it is  infeasible to directly employ general domain Visual Question Answering ({\sc vqa}) models for the medical domain.
Additionally, medical images annotation is a costly and time-consuming process.
To overcome these limitations, we propose a solution inspired by  self-supervised pretraining  of {\sc t}ransformer-style architectures for {\sc nlp}, {\sc v}ision and {\sc l}anguage tasks. 
Our method involves learning richer medical image and text semantic representations using {\sc m}asked {\sc l}anguage {\sc m}odeling ({\sc mlm}) with image features as the pretext task on a large medical image+caption dataset. The proposed solution achieves new state-of-the-art performance on two {\sc vqa}
datasets for radiology images -- {\sc vqa}-{\sc m}ed 2019 and {\sc vqa-rad}, outperforming even the ensemble models of previous best solutions. Moreover, our solution provides attention maps which help in model interpretability. The code is available at \url{https://github.com/VirajBagal/MMBERT}
\end{abstract}
\section{Introduction and Related Work}
\label{sec:intro}
Visual question answering ({\sc vqa}) on medical images aspires to build models that can answer diagnostically relevant natural language questions asked on medical images. It can provide valuable additional insights to medical professionals and can help the patients in the interpretation of their medical images.
However, supervised learning algorithms require large labeled datasets for effective performance and a major drawback of {\sc vqa} in the medical domain is the small size of existing  datasets ~\cite{vqa-rad,ImageCLEFVQA-Med2019,he2020pathvqa}. Since the annotations on medical images require the help of an expert, it is difficult to crowdsource and  annotation cost is high. This motivates the  usage of self-supervised pretraining methods.
\begin{figure}[t]
  \centering
  \begin{subfigure}{.4\columnwidth}
    \centering
    \includegraphics[width=\linewidth]{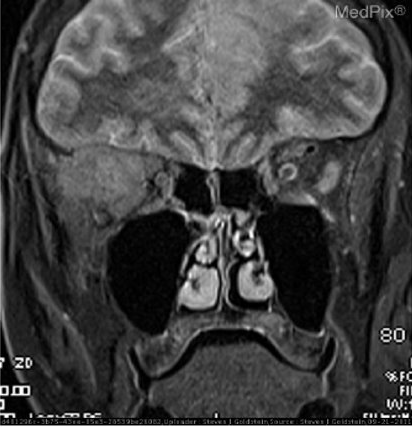}
  \end{subfigure}%
  \hspace{2em}
  \begin{subfigure}{.4\columnwidth}
    \centering
    \includegraphics[width=\linewidth]{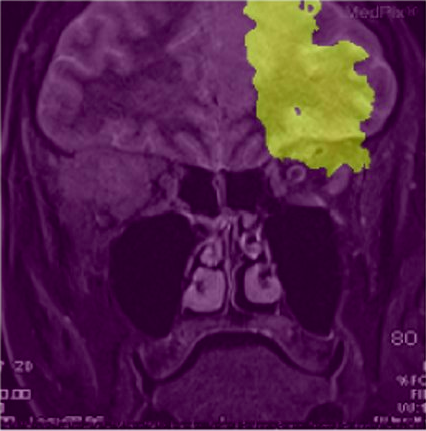}
  \end{subfigure}%
  \vspace*{3mm}
  \footnotesize{\normalsize \fontfamily{bch}\selectfont \textbf{Question}: What imaging modality was used?  \\
  \hskip-2.75cm \textbf{Answer}: MR-T2 Weighted}
  \vspace*{3mm}
  \caption{Example illustrating the attention map from our \mmbert model. For the given question, the model attends to grey matter, white matter and cerebrospinal fluid (CSF) and predicts the correct answer -- 'MR-T2 Weighted'.}
  \label{page1:figure}
  \vspace{-3mm}
\end{figure}

\begin{figure*}[t]
\begin{center}
  \includegraphics[width=0.8\linewidth, width=17.8cm, height=7cm]{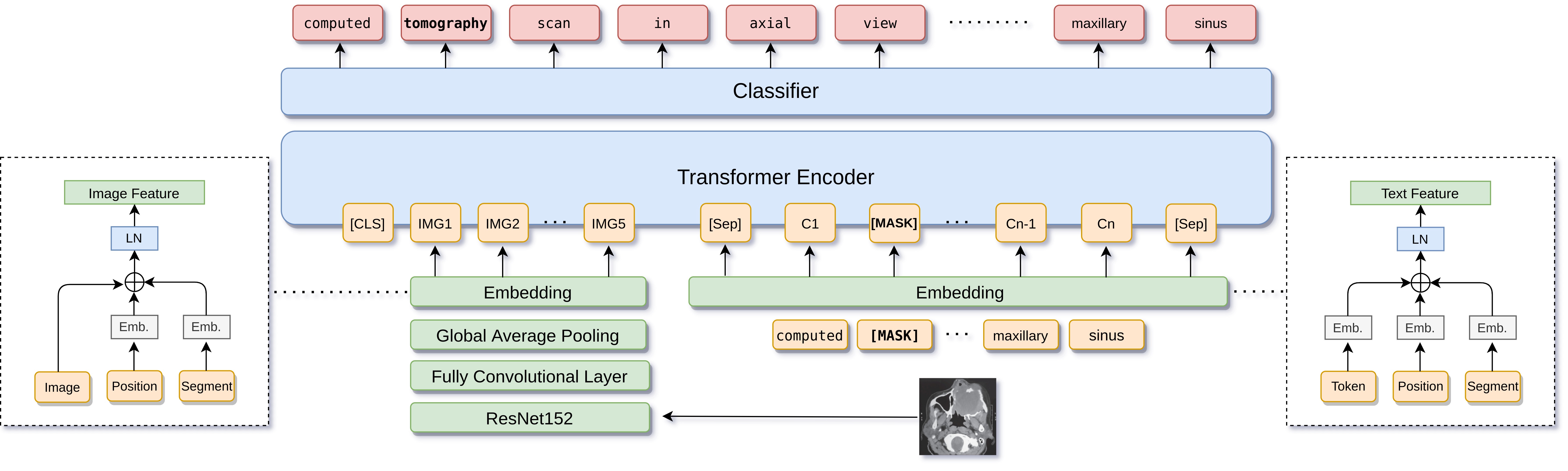}
\end{center}
  \caption{Model architecture for {\sc MLM} Pretraining task with image features on image caption data. To facilitate distinguishibility, we use embeddings of 0 and 1 as segment embeddings for image and text features respectively. We use embeddings of enumeration of image and text features as position embeddings for image and text respectively. The image features are extracted from ResNet152 and passed through an embedding layer. The caption is tokenized and the keywords are masked with [Mask] tokens. The text embeddings are obtained by combining input, position and segment embeddings. The final embedding is passed through a transformer encoder. The encoder outputs are then passed to a classifier which predicts the masked words.}
\label{fig:mlm}
\vspace{-3mm}
\end{figure*}



Self-supervised pretraining of {\sc bert}-like architectures  has
been proven quite effective in {\sc n}atural {\sc l}anguage
{\sc p}rocessing ({\sc nlp}) ~\cite{bert}, {\sc v}ision and {\sc l}anguage ~\cite{vilbert, chen2020uniter} space. The solution we propose - a {\sc m}ultimodal {\sc m}edical {\sc bert} ({\sc mmbert}) is inspired by these approaches. We first pretrain our {\sc mmbert}
model on a set of medical images and their corresponding
captions for the masked {\sc lm} task. Later this model is finetuned for the {\sc vqa} task.
\\
\indent
Yan et al.~\cite{winner} who are the winners of the \vqamed challenge, use a {\sc c}onvolutional {\sc n}eural {\sc n}etwork ({\sc cnn}) and {\sc bert} to extract image and question features respectively, followed by co-attention to fuse these features and a decoder to predict the answers. Ren et al.~\cite{cgmvqa} propose a model called {\sc cgmvqa} that uses a multimodal transformer architecture, similar to the proposed \mmbert. Zhan et al.\cite{zhan} use a conditional reasoning framework for medical {\sc vqa} on {\sc vqa-rad} dataset and they train a model separately for both the open-ended and closed-ended questions in the dataset.
\\
\indent
Although the aforementioned methods obtain effective results, they do not use existing large multimodal medical datasets to learn better image and text representations.
Our approach takes this into account and achieves new state-of-the-art performance on two medical {\sc vqa} datasets.
Our \mmbert, even with a single model for both the type of questions, yields better results than all the previous models on {\sc vqa-rad}. It also achieves a 5\% improvement in Accuracy over the previous state-of-the-art model on {\sc vqa}-{\sc m}ed 2019 dataset.
Moreover, our model provides attention maps and as shown in Fig \ref{page1:figure}, it focuses on correct region (grey and white matter difference) to predict the modality of the image.

\vspace{-0.2cm}

\section{Method}
\label{sec:method}

Transfer learning is quite popular in machine learning. However, a shift in image data distribution might result in sub-optimal performance when using pretrained weights from general domain. Moreover, there are changes in co-occurences of words in the medical text compared to the general domain text. These factors motivate the need for learning semantic representations of medical images and texts from scratch. Owing to the attention operation, we use Transformer encoder for learning effective representations.   

\subsection{Self-Attention}
\label{ssec:attention}

 Self-Attention allows attention to intra-modality and inter-modality features, thus enhancing the semantics of the intermediate representations. It involves mapping a query vector to the weighted addition of the value vectors where the weights are obtained by scaling the dot product of the query and the key vectors~\cite{vaswani2017attention}. The query, key and value vectors are represented together in the matrices $Q$, $K$ and $V$ respectively. The dot product of $Q$ and $K$ is scaled inversely by $\sqrt{d_k}$, where $d_k$ is the dimension of query and key vectors.  
\[Attention(Q,K,V) = softmax(\frac{QK^{T}}{\sqrt{d_{k}}})V\] 
\\
Instead of performing a single self-attention, the {\sc t}ransformer {\sc e}ncoder performs multiple self-attentions (multi-head attention) in parallel and concatenates the output. Multi-head attention provides better representations by attending to different representation subspaces at different positions.

\vspace{-0.3cm}

\begin{table*}[t]

    \begin{center}
    \small
    \begin{tabular}{lccccccccccccc}
    \hline
    \multicolumn{1}{c}{Method} &
    \multicolumn{1}{c}{Dedicated} &
    \multicolumn{2}{c}{Modality} & \multicolumn{2}{c}{Plane} & \multicolumn{2}{c}{Organ} & \multicolumn{2}{c}{Abnormality} & \multicolumn{2}{c}{Yes/No} & \multicolumn{2}{c}{Overall}\\
    \cline{3-14}
    \multicolumn{1}{c}{} & Models& Acc. & {\bleu} & Acc. & {\bleu} & Acc. & {\bleu} & Acc. & {\bleu} & Acc. & {\bleu} & Acc. & {\bleu}\\
    \hline
    {\sc vgg16}+{\sc bert}~\cite{winner}   &
    -&-&-&-&-&-&-&-&-&-&-&62.4&64.4\\
    {\sc cgmvqa}~\cite{cgmvqa}   &   
    \cmark&80.5&85.6&80.8&81.3&72.8&76.9&1.7&1.7&75.0&75.0&60.0&61.9\\
    {\sc cgmvqa} Ens.~\cite{cgmvqa} &  
    \cmark&81.9&\textbf{88.0}&\textbf{86.4}&\textbf{86.4}&\textbf{78.4}&79.7&4.40&7.60&78.1&78.1&64.0&65.9  \\
    \hline
    {\mmbert} {\sc g}eneral & \xmark&77.7&81.8& 82.4&82.9&73.6&76.6&5.20&6.70&85.9&85.9&62.4&64.2\\

    {\mmbert} {\sc np} & \cmark&80.6&85.6&81.6&82.1&71.2&74.4&4.30&5.70&78.1&78.1&60.2&62.7\\
    {\mmbert} {\sc e}xclusive      & \cmark&\textbf{83.3}&86.2&\textbf{86.4}&\textbf{86.4}&76.8&\textbf{80.7}&\textbf{14.0}&\textbf{16.0}&\textbf{87.5}&\textbf{87.5}&\textbf{67.2}&\textbf{69.0}\\

    \hline
    
    \end{tabular}
    \caption{Results on \vqamed dataset. Our method outperforms all previous methods that include methods with ensemble models in overall Accuracy and {\sc bleu} score. {\sc np} and {\sc e}ns. refer to non-pretrained and ensemble models respectively.}
    \label{tab:vqamed_results}
    \end{center}
    \vspace{-5mm}

\end{table*}

\begin{table}[t]
    \begin{center}
    \small
    \begin{tabular}{lc c ccc}
    \hline
    \multicolumn{1}{c}{Method}& 
    \multicolumn{1}{c}{Dedicated}&
    \multicolumn{3}{c}{Accuracy}&\\
    \cline{3-5}
    \multicolumn{1}{c}{} &
    Models&Open&Closed&Overall\\
    \hline
    {\sc mevf}+{\sc san}~\cite{nguyen2019overcoming}   & -& 40.7&74.1&60.8\\
    {\sc mevf}+{\sc ban}~\cite{nguyen2019overcoming}   &  -&43.9&75.1&62.7\\
    {\sc cr}~\cite{zhan}   &    \cmark&60.0&\textbf{79.3}&71.6\\
    \hline
    {\mmbert} {\sc g}eneral    &   \xmark&\textbf{63.1}&77.9&\textbf{72.0}\\
    \hline
    \end{tabular}
    \caption{Results on {\sc vqa-rad} dataset. Our method with single model for both open-ended and closed-ended question types outperforms all previous methods including methods with dedicated models for each question type in overall Accuracy.}
    \label{tab:vqarad_results}
    \end{center}
    \vspace{-5mm}

\end{table}

\subsection{Pretraining}
\label{ssec:pretraining}

A schematic of the \mmbert pretraining is shown in Figure~\ref{fig:mlm}.
For image features, similar to the {\sc cgmvqa}~\cite{cgmvqa} we use {\sc r}es{\sc n}et152~\cite{resnet} and extract features from different convolution layers. This helps in retaining information from different resolutions. We use {\sc bert} {\sc w}ord{\sc p}iece tokenizer~\cite{bert} for text tokenization. The sequence of 5 image features and the caption token embeddings together are provided as input to the {\sc bert}-like model. Unlike {\sc bert}\textsubscript{{\sc base}}~\cite{bert} our model has only 4 {\sc bert} {\sc l}ayers and a total of 12 attention heads.

We use masked language modeling with image features as the  pretraining task. In masked language modeling with image features, the task is to predict the original token in place of a [{\sc mask}] token with the usage of not only the accompanying text but also the image features. To ensure that the model learns to predict medical words from the context, we mask only medical keywords (provided with the dataset) from the captions and leave the common words untouched. 

\subsection{Finetuning}
\label{ssec:finetuning}

We load the model with weights from pretraining and finetune it further on the train split of the respective medical {\sc vqa} dataset. Instead of using [{\sc cls}] (Classification) token representation from the last layer of the {\sc t}ransformer, we average the representation of each token obtained from the last layer and further pass it through dense layers for classification.


\vspace{-0.25cm}
\section{Experiments and Results}
\label{sec:experiments}

\subsection{Data}
\label{ssec:datades}

{\sc r}adiology {\sc o}bjects in {\sc co}ntext (\roco)~\cite{roco} dataset contains over 81,000 radiology images with several medical imaging modalities.
For pretraining, we use all the images, their corresponding captions and use the keywords for masking. 
\vqamed~\cite{ImageCLEFVQA-Med2019} is a challenge dataset introduced as part of the Image{\sc clef}-{\sc vqa} {\sc m}ed 2019 challenge. It contains radiology images and has four main categories of questions: Modality, Plane, Organ system and Abnormality. All the samples having Yes/No as the ground truth are considered as Yes/No category. The dataset includes a training set of 3200 medical images with 12,792 Question-Answer ({\sc qa}) pairs, a validation set of 500 medical images with 2000 {\sc qa} pairs and a test set of 500 medical images with 500 {\sc qa} pairs. {\sc vqa-rad} has  315 images and 3515 questions of 11 types. 58\% of questions are close-ended while the rest are open-ended.   

\subsection{Experiments}
\label{ssec:experiments}

In our study, we primarily experiment with  three different settings for the \mmbert: (i) \underline{{\mmbert} {\sc g}eneral:} a model pretrained on \roco and finetuned on all samples in the train split of the respective {\sc vqa} dataset (ii) \underline{{\sc mmbert} {\sc e}xclusive:} an initial model pretrained on \roco, which is further finetuned separately for different question categories. For example in case of {\sc vqa}-{\sc m}ed 2019, we learn 5 different models, one for each question category and (iii) \underline{{\mmbert} {\sc n}on-{\sc p}retrained ({\sc np}):} Dedicated models for each question category but without pretraining on the \roco dataset.
At the time of the inference, for settings where there are dedicated models for each question category we first predict the question category using a {\sc bert}\textsubscript{{\sc base}} classifier.
\\
\indent
We train the models on a single {\sc nvidia} {\sc rtx} 2080{\sc t}i {\sc gpu}. For pretraining and finetuning, we resize all images to $224 \times 224$. We use image crop, rotation  and color jitter for augmentation. For pretraining, we optimize the loss using Adam optimizer~\cite{adam} with learning rate $2e-5$ and reduce the learning rate by a factor of 0.1 if the validation loss does not improve for 5 consecutive epochs. 
For finetuning, we use the Adam optimizer, learning rate of  $1e-4$ and reduce the learning rate by a factor of 0.1 if validation loss does not improve for 10 consecutive epochs. 



\begin{figure*}[ht]
\begin{center}
\small
\begin{tabular}{p{0.175\linewidth}p{0.175\linewidth}p{0.175\linewidth}p{0.175\linewidth}p{0.175\linewidth}}
    \includegraphics[width=3.25cm,height=3cm]{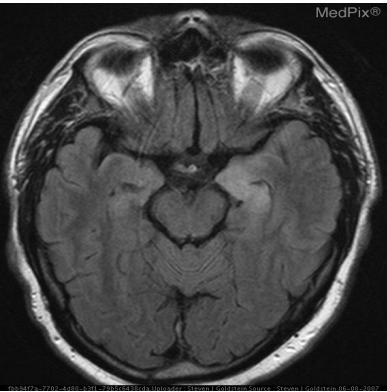} &
    \includegraphics[width=3.25cm,height=3cm]{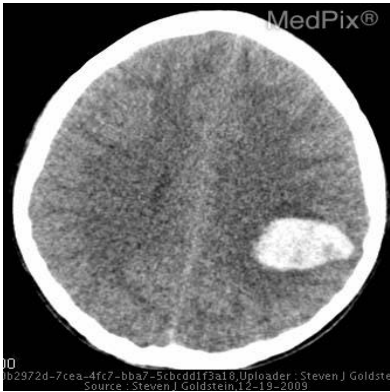} &
    \includegraphics[width=3.25cm,height=3cm]{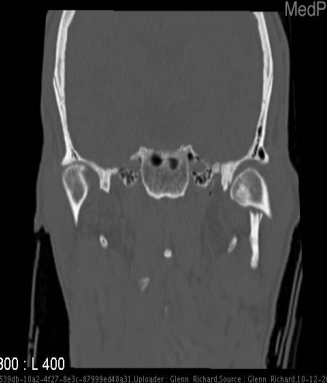} &
    \includegraphics[width=3.25cm,height=3cm]{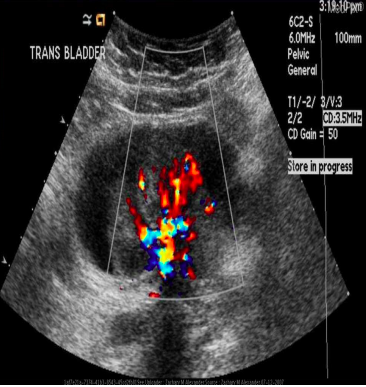} &
    \includegraphics[width=3.25cm,height=3cm]{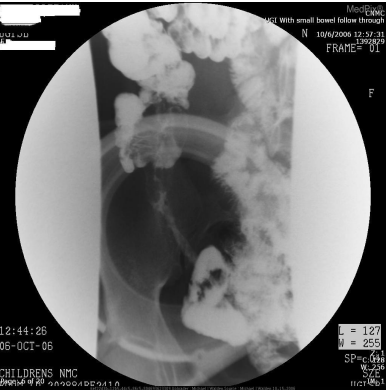}

    \\

  \includegraphics[width=3.25cm,height=3cm]{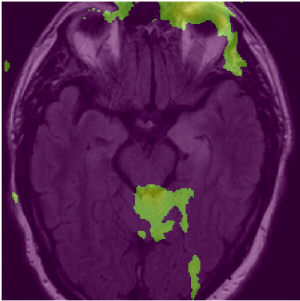} &
  \includegraphics[width=3.25cm,height=3cm]{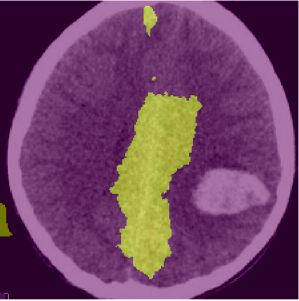} &
    \includegraphics[width=3.25cm,height=3cm]{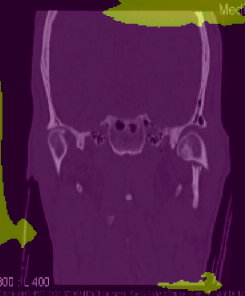} &
 \includegraphics[width=3.25cm,height=3cm]{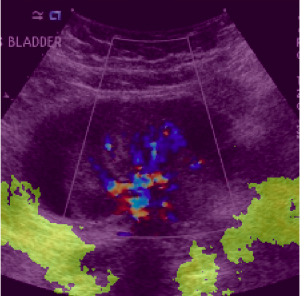} &
    \includegraphics[width=3.25cm,height=3cm]{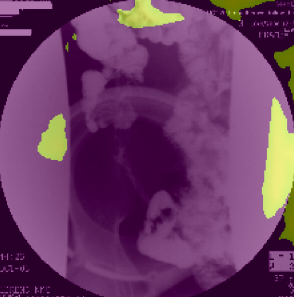}

    \\
    \footnotesize{\fontfamily{bch}\selectfont \textbf{C:} Organ} \par 
    \footnotesize{\fontfamily{bch}\selectfont \textbf{Q:} What organ is the image of? } \par \footnotesize{\fontfamily{bch}\selectfont \textbf{GT:} \color{blue}skull and contents} 
    \par \footnotesize{\fontfamily{bch}\selectfont \textbf{ME:} \color{green}skull and contents} 
    &
    
    \footnotesize{\fontfamily{bch}\selectfont \textbf{C:} Plane } \par 
    \footnotesize{\fontfamily{bch}\selectfont \textbf{Q:} What is the plane of the image? } \par  \footnotesize{\fontfamily{bch}\selectfont \textbf{GT:} \color{blue} axial} 
    \par \footnotesize{\fontfamily{bch}\selectfont \textbf{ME:} \color{green} axial} 
    &
   \footnotesize{\fontfamily{bch}\selectfont \textbf{C:} Yes/No} \par 
    \footnotesize{\fontfamily{bch}\selectfont \textbf{Q:} Is this an MRI image?} \par  \footnotesize{\fontfamily{bch}\selectfont \textbf{GT:} \color{blue} no} 
    \par \footnotesize{\fontfamily{bch}\selectfont \textbf{ME:} \color{green} no} 
   &
    \footnotesize{\fontfamily{bch}\selectfont \textbf{C:} Modality } \par 
  \footnotesize{\fontfamily{bch}\selectfont \textbf{Q:} What imaging method was used?} \par      
  \footnotesize{\fontfamily{bch}\selectfont \textbf{GT:} \color{blue} us-d - doppler ultrasound} 
    \par \footnotesize{\fontfamily{bch}\selectfont \textbf{ME:} \color{red} us - ultrasound} 
   &
   
   \footnotesize{\fontfamily{bch}\selectfont \textbf{C:} Abnormality} \par 
    \footnotesize{\fontfamily{bch}\selectfont \textbf{Q:} What is abnormal in the image?} \par  \footnotesize{\fontfamily{bch}\selectfont \textbf{GT:} \color{blue} crohn's disease} 
    \par \footnotesize{\fontfamily{bch}\selectfont \textbf{ME:} \color{red} fluoroscopic evaluation of small bowel in crohn's ileitis} 
   
\end{tabular}
\end{center}
\caption{ME and GT refers to {\mmbert} {\sc e}xclusive \& {\sc g}round {\sc t}ruth. The bottom row comprises attention maps for the corresponding top row images. In the Organ and Yes/No category, the model rightly attends to the bony part and soft tissue content to predict the right answer. In the Plane category, the model attends to the longitudinal fissure that is the key visual cue of the axial plane. The model fails to attend to the visual cue of doppler effect (colorful regions) in the Modality category. The Abnormality model surprisingly predicts a better answer than the ground truth by simultaneously predicting the modality, organ and abnormality.}
\label{fig:qualitative_results}
\vspace{-5mm}

\end{figure*}

\subsection{Results and Analysis}
\label{subsec: results}

We use Accuracy and {\sc B}i{\sc l}ingual {\sc e}valuation
{\sc u}nderstudy ({\bleu}) score to evaluate the {\sc vqa} performance. \bleu score is the percentage of uniformly weighted 4-grams in the predicted answer that are shared with the ground truth.
Table~\ref{tab:vqamed_results} reports  results on the \vqamed dataset.
Our {\mmbert} {\sc e}xclusive
achieves state-of-the-art results on the overall Accuracy and {\bleu} score, even surpassing {\sc cgmvqa} {\sc e}ns. which is an ensemble of 3 dedicated models for each category. Even our {\mmbert} {\sc g}eneral 
performs better than the {\sc cgmvqa} {\sc e}ns. on the Abnormality and Yas/No categories. Additionally, our {\mmbert} {\sc g}eneral outperforms single dedicated {\sc cgmvqa} models in all categories but Modality.  \\
\indent
 In the Organ category, {\mmbert} {\sc e}xclusive outperforms {\sc cgmvqa} {\sc e}ns. in \bleu but not in Accuracy. {\bleu} score is calculated by counting matching 1-gram in the predicted answer to the 1-gram in the ground truth. The comparison is made regardless of the order. This suggests that even though our model couldn't predict perfectly right answers, it could predict more answers close to the ground truth than the {\sc cgmvqa} {\sc e}ns. We find the opposite behaviour in the Modality category. 
When compared to {\mmbert} {\sc np}, we find that the pretraining increases the Accuracy and {\bleu} score by 7.2 and 9 points respectively.
\\
\indent
Table \ref{tab:vqarad_results} reports results on the {\sc vqa-rad} dataset. {\mmbert} {\sc g}eneral, which  is a single model for both the question types in the dataset, outperforms the existing approaches including the ones which have a dedicated model for each question type.

\subsection{Qualitative Analysis}
\label{subsec: qual_analysis}

Fig. ~\ref{fig:qualitative_results} shows the category-wise qualitative results from {\mmbert} {\sc e}xclusive. The top row comprises the original images while the bottom row comprises the attention maps obtained from our model. The attention maps highlight the regions in the image which contribute the most to the prediction. In the Organ and Yes/No category, the model rightly attends to the skull (the bony part) and its contents (brain tissues) to predict the right answer. In the Plane category, the model attends to the longitudinal fissure which is the key visual cue in identifying the axial plane as it separates the brain into Right and Left hemispheres.
In the Modality category, the model attends to the soft tissue and fluid part of the image and is able to correctly predict that it is an ultrasound image. However, it fails to attend to the visual cue of the Doppler (the colour region) and hence fails to correctly answer. Surprisingly, in the case of Abnormality category, for the attention map
resulting for the "{\sc f}luoroscopic evaluation of small bowel
in {\sc c}rohn's ileitis" prediction, our model predicts a better answer than the ground truth. Here, it is simultaneously predicting the modality (fluoroscopy), organ (bowel) and the abnormality (Crohn's ileitis).\\
\indent
Medical experts find it difficult to make a correct diagnosis of abnormalities from a single image.
They often resort to multiple sections (slices), planes, and other evidences. On closely analyzing our results we see that our model predicts abnormalities which could have also been a differential diagnosis for a human expert. However, our quantitative evaluation protocol does not take this into consideration.

\section{Conclusion}

In this work, we prospose to pretrain {\sc m}ultimodal {\sc m}edical {\sc bert} (\mmbert) on {\sc roco} dataset with masked language modeling using image features for medical {\sc vqa}. We finetune it on {\sc vqa-rad} and {\sc vqa}-{\sc m}ed 2019 datasets and achieve new state-of-the-art results on these datasets. Moreover, qualitative results show that our models can rightly attend to the image regions for prediction.

\label{sec:conc}

\section{Compliance with Ethical Standards}
\label{sec:ethics}
This research study was conducted retrospectively using human subject data made available in open access by O. Pelka et al.\cite{roco}, Asma Ben Abacha et al.\cite{ImageCLEFVQA-Med2019} and Lau et al.\cite{vqa-rad}. Ethical approval was not required as confirmed by the license attached with the open access data.
\section{Acknowledgments}
\label{sec:acknowledgments}
No funding was received for conducting this study. The authors have no relevant financial or non-financial interests to disclose.

{\small
\bibliographystyle{IEEEbib}
\bibliography{refs}
}
\end{document}